\documentclass[sigconf,pagenumber,hyperref,table]{acmart}
\usepackage{algorithm}
\usepackage{algorithmicx}

\usepackage{marvosym}
\usepackage{ifsym}

\usepackage[noend]{algpseudocode}  
\usepackage{pdfpages}
\usepackage{amsmath}  

\AtBeginDocument{%
  }

\copyrightyear{2023} 
\acmYear{2023}  
\setcopyright{acmlicensed}\acmConference[MM '23]{Proceedings of the 31st
ACM International Conference on Multimedia}{October 29-November 3,
2023}{Ottawa, ON, Canada}
\acmBooktitle{Proceedings of the 31st ACM International Conference on
Multimedia (MM '23), October 29-November 3, 2023, Ottawa, ON, Canada}
\acmPrice{15.00}
\acmDOI{10.1145/3581783.3612145}
\acmISBN{979-8-4007-0108-5/23/10}

\begin{document}

\title{CLE Diffusion: Controllable Light Enhancement Diffusion Model}

\begin{teaserfigure}
    \centering
\includegraphics[width=\textwidth]{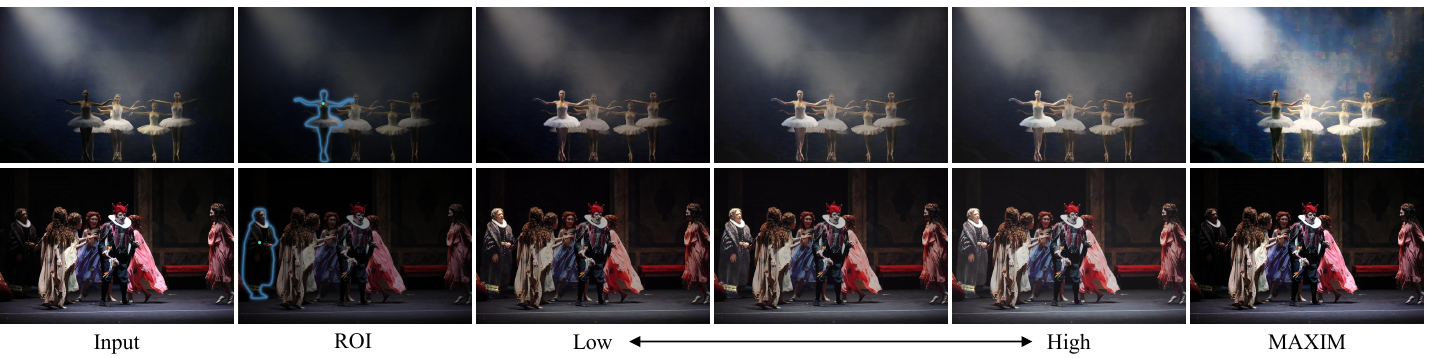}
\caption{
CLE Diffusion enables users to select regions of interest(ROI) with a simple click and adjust the degree of brightness enhancement as desired, while MAXIM~\cite{tu2022maxim} is limited to homogeneously enhancing images to a pre-defined level of brightness.
}
\label{fig:region_control}
\end{teaserfigure}

\author{Yuyang Yin}
\affiliation{%
  \institution{Institute of Information Science, Beijing Jiaotong University}
 \institution{ Beijing Key Laboratory of Advanced Information Science and Network Technology}
  \country{China}
}

\author{Dejia Xu}
\affiliation{%
  \institution{VITA Group}
  \institution{University of Texas at Austin}
  \country{USA}
}

\author{Chuangchuang Tan}
\affiliation{%
  \institution{Institute of Information Science, Beijing Jiaotong University}
 \institution{ Beijing Key Laboratory of Advanced Information Science and Network Technology}
  \country{China}
}
  
\author{Ping Liu}
\affiliation{%
  \institution{Center for Frontier AI Research, IHPC, A*STAR}
  \country{Singapore}
}

\author{Yao Zhao}
\affiliation{%
  \institution{Institute of Information Science, Beijing Jiaotong University}
 \institution{ Beijing Key Laboratory of Advanced Information Science and Network Technology}
  \country{China}
}

\author{Yunchao Wei {\Letter}}
\affiliation{%
  \institution{Institute of Information Science, Beijing Jiaotong University}
 \institution{ Beijing Key Laboratory of Advanced Information Science and Network Technology}
  \country{China}
}
 \renewcommand{\shortauthors}{Yuyang Yin et al.}

\begin{abstract}
Low light enhancement has gained increasing importance with the rapid development of visual creation and editing. 
However, most existing enhancement algorithms are designed to homogeneously increase the brightness of images to a pre-defined extent, limiting the user experience.
To address this issue, we propose Controllable Light Enhancement Diffusion Model, dubbed \textbf{CLE Diffusion}, a novel diffusion framework to provide users with rich controllability. 
Built with a conditional diffusion model, we introduce an illumination embedding to let users control their desired brightness level.
Additionally, we incorporate the Segment-Anything Model (SAM) to enable user-friendly region controllability, where users can click on objects to specify the regions they wish to enhance.
Extensive experiments demonstrate that CLE Diffusion achieves competitive performance regarding quantitative metrics, qualitative results, and versatile controllability.
Project page: \url{https://yuyangyin.github.io/CLEDiffusion/}
\end{abstract}



 \maketitle

\section{Introduction}

The low-light capturing conditions can stem from various factors, ranging from environmental causes like inadequate illumination to technical reasons such as sub-optimal ISO settings.
Consequently, these images are often regarded as less visually appealing and present new difficulties for downstream tasks.
Remedying such degradation has been a critical area of vision research for many years, leading to the development of diverse approaches and solutions for low-light enhancement.

Enhancing low-light images tackles an ill-posed problem due to the lack of large publicly available datasets with paired images, which makes it difficult to create a universal solution for quick and easy enhancement.
As a result, this challenging problem has remained the focus of numerous researchers over the past few decades.
One approach to addressing this issue is using histogram equalization (HE) methods~\cite{HE1, HE2} to adjust the contrast. These methods aim to stretch the dynamic range of low-light images, but they often result in unwanted illumination in more intricate scenes.
Another line of research involves the implementation of the Retinex theory~\cite{Land77theretinex}, which aims to decompose low-light images into two distinct layers, reflectance, and illumination. Various image filters~\cite{Multi_scale_retinex,lee2013amsr,NPE} and manually designed priors~\cite{Fu_2016_WTV,Guo_2017_Lime} have been utilized to enhance the decomposition process.

With the recent development of deep learning, data-driven approaches for low-light enhancement have gained significant interest in recent years.  These approaches utilize large-scale datasets to restore normal-light images from complex degradations. However, many of these methods~\cite{ren2019low,shen2017msr,wang2019underexposed,DRD} require paired low-light and normal-light images pixel-aligned to achieve optimal results. Recently, researchers have made significant progress by utilizing unpaired data through adversarial learning~\cite{jiang2021enlightengan,xu2022recoro} and more advanced networks~\cite{fan2022half,wang2021low,yang2020fidelity,guo2020zero} for enhancing low-light images. 

However, adversarial learning is prone to optimization instability~\cite{arjovsky2017wasserstein,gulrajani2017improved} and mode collapse issues~\cite{metz2016unrolled,ravuri2019classification}, making them hard to scale up. 
The recent success of denoising diffusion models~\cite{ho2020denoising,song2020denoising,rombach2022high,nichol2021glide,ramesh2022hierarchical,saharia2022photorealistic} in image generation reveals their outstanding capacity. This has also attracted researchers to study their abilities for image restoration tasks~\cite{kanizo2013palette,saharia2022image,whang2022deblurring}. These attempts have proven that denoising diffusion models can model natural image distributions better than existing methods. Despite the exciting progress, there has been no effort to explore diffusion models' ability to restore low-light images.

Moreover, most existing approaches are implicitly designed to enhance low-light images in a deterministic way, ignoring the ill-posed nature of low-light enhancement. 
While learning a one-to-one mapping between low-light and normal-light images might provide visually pleasing results, this design limits the model's flexibility since they assume a pre-defined well-lit brightness exists. Although the claim might hold for certain natural images, the definition of well-lit remains highly subjective. On the contrary, the presumed well-lit brightness is extracted through the datasets~\cite{Chen2018Retinex,liu2021benchmarking,fivek} for most data-driven methods. Despite the great effort in collecting aligned datasets~\cite{liu2021benchmarking,Chen2018Retinex,fivek}, low-light and normal-light image pairs remain scarce compared to large-scale datasets like ImageNet~\cite{deng2009imagenet} and LAION~\cite{schuhmann2022laion}.
As a result, the learned mapping is biased toward the available samples, and an ability to provide a diverse set of plausible reconstructions is greatly desired. While ReCoRo~\cite{xu2022recoro} provides controllability of brightness level, the model assigns the brightest illumination using the training set pairs. This brings brightness inconsistency across the images, leading to unstabilized optimization and confusing user experience. Moreover, ReCoRo has little extrapolation ability to generate brighter results than the ground truth images in the dataset.

On the other hand, most previous methods learn to enlighten images in a global homogeneous way, making the models incapable of dealing with intricate lighting situations where both under-exposed and over-exposed regions coexist. Though these methods may improve the visibility of under-exposed regions, they often result in over-exposed regions becoming overly emphasized since the models fail to address them adaptively.
To this end, users are used to specifying regions of interest manually so that the desired edits can be performed locally without retouching the other regions.
However, this requirement of precise user specification can be tedious for smartphone end-users since finger drawings inevitably come with unintentional noise. ReCoRo~\cite{xu2022recoro} studies this setting for the first time and develops a robust enhancement model to work with imprecise user inputs by baking in domain-specific augmentations. However, their augmentations are specially designed for portrait images and require further tuning for masks of other image classes.

To overcome the above-mentioned issues, we propose a novel denoising diffusion framework, dubbed \textbf{CLE Diffusion}, that performs \underline{C}ontrollable \underline{L}ight \underline{E}nhancement via iterative refinement. 
Conditioned on a unified illumination embedding, our diffusion model learns to enhance the low-light images towards a target brightness level specified by users. Our brightness level, different from ReCoRo's~\cite{xu2022recoro}, is represented using the average pixel intensities of the image and thus is consistent throughout the dataset. This avoids the wild assumption that a perfect well-lit illumination level exists.  We further condition the denoising diffusion process with the low-light image features to ease the optimization. Alongside the original low-light image, we prepare a normalized color map and a signal-to-noise ratio (SNR) map to reduce the burden of the enhancement module.
Additionally, we include a binary mask as an extra input to support localized edits, letting the users freely specify the regions of interest (ROI). Armed with the Segment-Anything Model (SAM)~\cite{sam}, our framework is capable of user-friendly region-controllable enhancement via diverse simple promptings, such as points and boxes, alleviating the requirement for precise user specification in practice.

Our major contributions can be summarized as follows,
\begin{itemize}
    \item We propose a novel diffusion framework, dubbed \textbf{CLE Diffusion}, for \underline{C}ontrollable \underline{L}ight \underline{E}nhancement via iterative refinement. To the best of our knowledge, CLE Diffusion is the first attempt to study controllable light enhancement using diffusion models.
    \item Our framework's controllability allows users to specify desired brightness levels and the regions of interest easily. Using a unified illumination embedding, our conditional diffusion model provides seamless and consistent control over brightness levels. Moreover, we facilitate user-friendly region control by employing the SAM model, which enables users to click on the image to specify regions for enhancement.
    \item As shown in extensive experiments, our CLE Diffusion demonstrates competitive performance in terms of quantitative, qualitative, and versatile controllability.
\end{itemize}

\section{Related Works}

\subsection{Traditional Light Enhancement Methods}

Many image priors have found their use for traditional single-image low-light enhancement. Some approaches implement local and global histogram equalization ~\cite{HE1,HE2} to increase the contrast of the input image. Some other solutions ~\cite{dehaze1,dehaze2} consider the low-light images as inverted haze images and use dehazing methods on the inverted input image. Another popular line of work is based on the Retinex theory~\cite{Land77theretinex}, which separates the image into illumination and reflectance layers and performs simple transformations on top of them for the desired effects. SRIE~\cite{fu2015probabilistic} estimates both layers simultaneously using a weighted variational model, while LIME~\cite{Guo_2017_Lime} refines the illumination layer estimate and uses the decomposed reflection layer as the final enhanced result.
For noise suppression, JED~\cite{ren2018joint} has made progress by utilizing sequential decomposition. However, hand-crafted models used in these methods require careful parameter tuning and have limited model capacity.

\begin{figure*}  %

      \centering  %

      \includegraphics[width=0.85\linewidth]{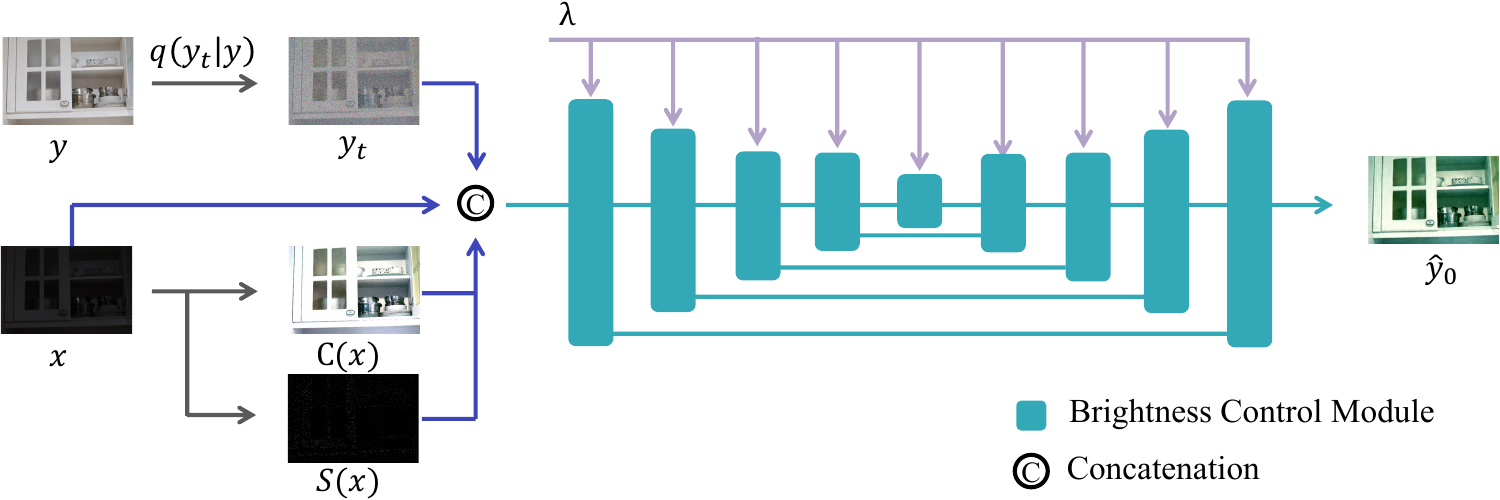} 

      \caption{The overall framework of our CLE Diffusion. During training, we randomly sample a pair of low-light image $x$ and normal-light image $y$. We then construct $y_t$, $C(x)$, and $S(x)$ as additional inputs to the diffusion model. Brightness level $\lambda$ is injected into the Brightness Control Modules to enable seamless and consistent brightness control. Alongside $\mathcal{L}_\text{simple}$, we introduce auxiliary losses on the denoised estimate $\hat{y_0}$ to provide better supervision for the model.}

      \label{fig:framework} 

\end{figure*}

\subsection{Learning-based Light Enhancement Methods}

\looseness=-1
In recent years, various learning-based light enhancement methods have been introduced thanks to the rapid development of deep learning.
LLNet and S-LLNet~\cite{LLNet} utilize deep autoencoder-based approaches for contrast enhancement and denoising. They are trained using data synthesized with random Gamma corrections and Gaussian noise. Based on the Retinex theory, Retinex-Net~\cite{Chen2018Retinex} assumes that paired aligned images share the same reflectance but have different illuminations. The authors collect a paired low-light and normal-light data set, which paves the way for the training of larger models.
Since collecting well-aligned paired datasets~\cite{liu2021benchmarking,fivek} requires hard work, great effort has been put into searching for methods without the need for paired supervision. EnlightenGAN ~\cite{jiang2021enlightengan} utilizes a generative adversarial network framework to learn a powerful generator without using unpaired data.
Zero-DCE~\cite{guo2020zero} learns light enhancement through image-specific curve estimation. CERL ~\cite{chen2022cerl} builds upon EnlightenGAN and incorporates plug-and-play noise suppression.
In addition, many works investigate the performance of different architectures, including Recursive Band Network~\cite{yang2020fidelity}, Signal-to-Noise-Ratio (SNR) Prior-aware Transformer~\cite{xu2022snr}, Multi-axis MLP~\cite{tu2022maxim}, Normalizing Flow~\cite{wang2021low}, and Half Wavelet Network~\cite{fan2022half}.
These existing methods generally enhance the images to a pre-defined brightness level learned from the training dataset and avoid the challenging real-world case of region controlling for light enhancement.

A controllable GAN network named ReCoRo~\cite{xu2022recoro} is the closest baseline to our approach. It allows users to specify the areas and levels of enhancement. However, the authors assign the brightest illumination level for each image using its normal-light version from the training set. This inconsistency of brightness level across images unstabilizes network optimization and disturbs the user experience at inference time. ReCoRo~\cite{xu2022recoro} works with imprecise region controls by baking in augmentations. Effective as it is for portrait images, this method requires re-training for novel object classes.
In contrast, our CLE Diffusion framework produces visually pleasing results via a conditional diffusion framework.
Our unified brightness level also allows for seamless and consistent brightness control. Moreover, our framework offers a user-friendly experience with region control accessible through a single click.

\subsection{Diffusion Model}

\looseness=-1
Denoising diffusion model~\cite{sohl2015deep} is a deep generative model synthesizing data through an iterative denoising process. Diffusion models consist of a forward process that distorts clean images with noise and a reverse process that learns to reconstruct clean images. They have demonstrated outstanding image generation~\cite{song2019generative,ho2020denoising} capability with the help of various improvements in architecture design~\cite{dhariwal2021diffusion,rombach2022high}, sampling guidance~\cite{ho2022classifier}, and inference cost~\cite{salimans2022progressive, rombach2022high, vahdat2021score}. Equipped with large-scale image-pair datasets, many works scaled up the model ~\cite{rombach2022high, nichol2021glide, ramesh2022hierarchical, saharia2022photorealistic} to billions of parameters to work with the challenging text-to-image generation. Additionally, denoising diffusion models have succeeded in various high-level computer vision tasks, including 3D generation~\cite{poole2022dreamfusion,singer2023text,xu2022neurallift,gu2023nerfdiff,watson2022novel}, object detection~\cite{chen2022diffusiondet} and depth estimation~\cite{saxena2023monocular}.
Meanwhile, various image restoration tasks are also studied through  diffusion models, including super resolution~\cite{saharia2022image}, 
 image deblurring~\cite{whang2022deblurring}, adverse weather condition~\cite{ozdenizci2023restoring}, and image to image translation~\cite{saharia2022palette}.
Our CLE Diffusion lies in the conditional denoising diffusion model category and makes the first attempt to study its effectiveness in controllable light enhancement.

\section{Method}

We propose a novel Controllable Light Enhancement (CLE) Diffusion framework, which adopts a conditional diffusion model to enhance any region in low-light images to any brightness level. As shown in Fig.~\ref{fig:framework},  we design domain-specific conditioning information and loss functions tailored to our needs. Additionally, we incorporate a Brightness Control Module to enable controllable light enhancement.
To further improve usability, we support region controllability by including a binary mask as input and leveraging the Segment-Anything Model (SAM)~\cite{sam}. This allows for a user-friendly interface where users can easily click on an image to specify the regions to enhance.

\subsection{Preliminary of Diffusion Model}
Diffusion model~\cite{sohl2015deep} is a type of generative model that uses iterative refinement to generate data. The widely used DDPM model~\cite{ho2020denoising} consists of a forward and a reverse process. The forward process gradually adds Gaussian noise to a clean input image, formulated as $q(y_t|y_{t-1})=\mathcal{N}(y_t;\sqrt{\beta_t}y_{t-1},(1-\beta_t)I)$.
Furthermore, the intermediate steps can be marginalized out to characterize the distribution into $q(y_{t}|y_{0}) = \mathcal{N}(\sqrt{\bar{\alpha_t}}y_0, (1-\bar{\alpha_t}) I)$, where $\alpha_t=1-\beta_t$ and $\bar{\alpha_t}=\prod_{i=0}^t\alpha_i$. The sequence of $\beta_t$ is carefully designed so that the noisy images will converge to pure Gaussian random noise $\mathcal{N}(0, I)$ when $t$ reaches the end of the forward process.

In the reverse process of the DDPM model, a neural network is used to denoise the data samples. The process can be formulated as 
$p_\theta(y_{t-1}|y_t) = \mathcal{N}(\hat{y_0},\frac{1-\bar{\alpha_{t-1}}}{1-\bar{\alpha_{t}}}\beta_t),$
where $\hat{y_0} =\frac{1}{\sqrt{\alpha_t}}(y_t-\frac{\beta_t}{\sqrt{1- \bar{\alpha}_t}}\epsilon_{\theta}(y_t,t)).$
$\epsilon_{\theta}$ is implemented with a  U-Net~\cite{ronneberger2015u} model to estimate the noise component from a noisy image. 
During inference time, we sample $y_T \sim \mathcal{N}(0, I)$ and gradually reduce the noise level until we reach a clean image $y_0$.
To accelerate sampling, DDIM~\cite{song2020denoising} presents a deterministic sampling approach as follows,
\begin{equation}
\begin{aligned}
y_{t-1}= & \sqrt{\bar{\alpha}_{t-1}}\left(\frac{y_{t}-\sqrt{1-\bar{\alpha}_{t}} \cdot \boldsymbol{\epsilon}_{\theta}\left(y_{t}, t\right)}{\sqrt{\bar{\alpha}_{t}}}\right) \\
& +\sqrt{1-\bar{\alpha}_{t-1}} \cdot \boldsymbol{\epsilon}_{\theta}\left(y_{t}, t\right).
\end{aligned}
\end{equation}

Diffusion models are usually trained via optimizing the negative log-likelihood loss function, which is further simplified~\cite{ho2020denoising,kingma2021variational} as follows:
\begin{equation}
\mathcal{L}_\text{simple}=\mathbb{E}_{\mathbf{y}_{0}, t, \epsilon}\left[\left\|\epsilon-\epsilon_{\theta}\left(\sqrt{\bar{a}_{t}} \mathbf{y}_{0}+\sqrt{1-\bar{a}_{t}} \epsilon,t\right)\right\|^{2}\right]
\label{eq:simple}
\end{equation}

\subsection{Controllable Light Enhancement Diffusion Model}  %
Specifically for low-light enhancement, we need to generate coherent normal-light images $y$ that share the content with the input low-light image $x$. 
Instead of learning the one-to-one mapping between the two domains, we are interested in approximating the conditional distribution $p(y|x)$ using the available paired image samples from the dataset.
Similar to existing attempts to utilize diffusion model for image restoration~\cite{saharia2022palette,saharia2022image}, we implement our CLE Diffusion by adapting the original DDPM~\cite{ho2020denoising} model to accept additional condition information.
While the forward process remains as simple as distorting the image with some carefully attenuated Gaussian noise, the U-net in the reverse process takes $x$ as an additional input.

Empirically, we observe simple concatenation of low-light image $x$ and diffused element $y_t$ leads to unstable training for complex scenes. This is partially due to the intricate noise and the diverse lighting in low-light environments. 
To this end, we further parameterize the conditional diffusion process with two additional priors.

Our first motivation comes from the severe color distortion in low-light images. As discussed in earlier works~\cite{xu2022recoro, liu2021benchmarking}, unnatural color shifts are often observed when enhancing low-light images. To this end, we implement a color map to reduce color distortion by normalizing the range of three color channels in input images. Specifically, an input image $x$ can be decomposed into three channels:
\begin{equation}
x=[x_r,x_g,x_b],
\end{equation}    
 where $x_r$ means the red channel of the image, $x_g$ means the green channel of the image, $x_b$ means the blue channel of the image.
 We then extract the maximum pixel value for the three channels respectively:
 \begin{equation}
x_{max}=[x_{r,max},x_{g,max},x_{b,max}],
\end{equation} 
where for example, $x_{r,max}$ means the maximum value of pixels on the red channel. Overall, the color map can be formulated as follows:
\begin{equation}
C(x)=\frac{x}{x_{max}}=[\frac{x_r}{x_{r, max}},\frac{x_g}{x_{g,max}},\frac{x_b}{x_{b,max}}],
\label{eq:cloromap}
\end{equation}

Another main challenge for enhancing low-light images lies in the inevitable noise in low-light conditions. Numerous researchers~\cite{ren2019low,Li_2017_SRRM,Chen2018Retinex} have made various attempts to model the noise formulation. More recently, Xu~\textit{et al.}~\cite{xu2022snr} adopted an SNR-aware transformer for effective low-light enhancement. Specifically, the SNR map is used to bring spatial attention to the low signal-to-noise-ratio region. The SNR map can be obtained as follows,
\begin{equation}
S(x)=\frac{F(x)}{\vert x - F(x) + \epsilon \vert},
\end{equation}  
where $\epsilon$ is included for numerical stability and $F$ is a low-pass filter  implemented as a Gaussian blur in our experiments. We consider the high-frequency component in the image to be noise and calculate the ratio between the original image and noise directly.

In each training step, we randomly sample a pair of low-light image $x$ and its corresponding normal-light image $y$. We then prepare the color map $C(x)$, the SNR map $S(x)$ and a noisy image $y_t=\sqrt{\bar{a}_{t}} {y}+\sqrt{1-\bar{a}_{t}} \epsilon$. They are concatenated with low-light image $x$ as the input of our diffusion model. 
Consequently, Eq.~\ref{eq:simple} can be extended as follows,
\begin{equation}
\begin{aligned}
\mathcal{L}_\text{simple}=\mathbb{E}_{{y}, t, \epsilon}\left[\left\|\epsilon-   
 \epsilon_{\theta}\left(\sqrt{\bar{a}_{t}} {y}+\sqrt{1-\bar{a}_{t}} \epsilon, x,C(x),S(x),t\right)\right\|^{2}\right]
\end{aligned}
\end{equation}

\begin{figure}
\centering
\includegraphics[width=0.8\columnwidth,keepaspectratio]{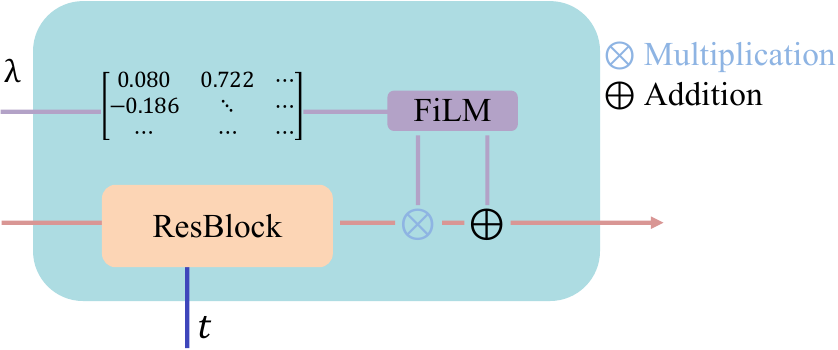}\\
\caption{Architecture of our Brightness Control Module.}
\label{fig:block}
\end{figure}

\subsection{Brightness Control Module}

\looseness=-1
To emphasize the effectiveness of conditioning information, we adopt a classifier-free guidance~\cite{ho2022classifier} approach.
This involves jointly training a conditional and an unconditional diffusion model, without the need for a classifier on noisy images as in the classifier guidance technique~\cite{dhariwal2021diffusion}. Instead, the classifier-free guidance approach allows for a trade-off between sample quality and diversity by adjusting the weighting of conditional sampling and unconditional sampling during inference.
For CLE Diffusion, we treat the brightness level of images as our "class" labels. It is important to note that our "class" in this case is continuous, allowing for seamless interpolation and continuous adjustment of the target brightness levels.

\looseness=-1
We first extract the vanilla brightness level $\lambda$ of normal-light images by calculating the average pixel value. Then we encode the average value using a random orthogonal matrix into the illumination embedding.
Specifically, we uniformly establish several discretized values in $[0, 1]$ as anchors. Given a $n \times n$ encoding matrix, the $d$-th value is mapped to the $d$-th column of the $n$-dimensional random orthogonal matrix.
We adopt bilinear interpolations of the two nearby columns for the intermediate values.
The illumination embedding is further embedded into the U-net using our Brightness Control Module. As shown in Fig.~\ref{fig:block}, we adopt a FiLM layer~\cite{perez2018film} to learn feature-wise affine transformation based on the illumination embedding. Then the modulated feature is split by half along the channel axis, with one copy being multiplied by the feature map and the other added to the feature map.

\looseness=-1
We train the conditional diffusion model $\epsilon_{\theta}(y_t,x,C(x),S(x)|\lambda)$ together with an unconditional model $\epsilon_{\theta}(y_t,x,C(x),S(x)|0)$. To be more specific, for the unconditional model, we input a zero embedding with the same shape as the illumination embedding. During the training process, we randomly drop the conditioning by allowing $2\%$ of the total iterations to train the unconditional model.

As shown in Algo.~\ref{art:sampling}, the sampling process is implemented with DDIM~\cite{song2020denoising} sampler. Firstly, we sample $y_T \sim \mathcal{N}(0, I)$ . Subsequently, we estimate two noise maps, one from the conditional model and the other from the unconditional model, and apply a weighted average of the two noise estimates.
The guidance scale $s$ is used to regulate the influence of the conditioning signal, where a larger scale produces results more aligned with the controlling signal, while a smaller scale produces less connected results.

\begin{algorithm}[t]  
  \caption{CLE Diffusion sampling}  
  \label{art:sampling}
  \begin{algorithmic} %
    \State \textbf{Input} 
    $y_T\sim\mathcal{N}(0,1)$,low-light image $x$,color map $C(x)$, SNR map $S(x)$, brightness level $\lambda$, and number of implicit sampling steps $T$  
      \For {$t=T,...,1$} 
      \State $
      \begin{aligned}
      e_{t-1}& = s\cdot \epsilon_{\theta}(y_t,x,C(x),S(x)|\lambda) \\ 
      & +(1-s)\cdot \epsilon_{\theta}(y_t,x,C(x),S(x)|0)
      \end{aligned}
      $
      \State $\hat{y_0} = \frac{y_{t}-\sqrt{1-\overline{\alpha_t}}\cdot e_{t-1}}{\sqrt{\overline{\alpha_t}}}$
      \State $y_{t-1}=\sqrt{\overline{\alpha_{t-1}}}\hat{y_0}+\sqrt{1-\overline{\alpha_{t-1}}}\cdot e_{t-1}$
        \EndFor  
\State \textbf{Output} $y_0$
  \end{algorithmic}  
\end{algorithm}

\subsection{Regional Controllability}
Users may prioritize increasing the brightness of specific regions of interest over globally illuminating the entire image, especially when dealing with complex lighting conditions.
To address this need, we introduce region controllability to our CLE Diffusion method, referred to as \textbf{Mask-CLE Diffusion}.

We incorporate a binary mask $M$ into our diffusion model by concatenating the mask with the original inputs. To accommodate this requirement, we created synthetic training data by randomly sampling free-form masks~\cite{suvorov2022resolution} with feathered boundaries. The target images are generated by alpha blending the low-light and normal-light images from existing low-light datasets~\cite{Chen2018Retinex,fivek}. 

Segment-Anything Model (SAM)~\cite{sam} is a large vision transformer trained on 11 million images to produce high-quality object masks from input prompts such as points or boxes.
With the aid of SAM~\cite{sam}, obtaining precise object masks in low-light conditions becomes a user-friendly process. Users can select their desired region with just one click. Mask-CLE Diffusion subsequently generates controllable light enhancement images via mixing the results from $\epsilon_{\theta}(y_t,x,C(x),S(x),M|\lambda)$ and $\epsilon_{\theta}(y_t,x,C(x),S(x),M|0)$.
For tiny objects such as human hair, SAM masks are not accurate enough. In such cases, we can still use the SAM mask as an initialization for downstream models. In our experiments, we utilize MatteFormer~\cite{park2022matteformer} to automatically construct detailed alpha mattes for human matting. The trimaps used for MatteFormer are constructed using dilate and erode operations.

\subsection{Auxiliary Loss Functions}
\looseness=-1
 Although an ideal optimum for the $\mathcal{L}_\text{simple}$ will be able to approximate the $p(y|x)$ distribution from the available paired dataset, in practice, we observe that the model often ends up generating color distortion and unprecedented noise. To improve the convergence of our diffusion model, we introduce auxiliary losses to provide direct supervision on the estimated denoised images $\hat{y_0}$.

\paragraph{\textbf{Brightness Loss}}
To maintain the same brightness level between the enhanced images and ground truth, we utilize brightness intensity loss. We use average pixel intensities to supervise the brightness information as follows, 
\begin{equation}
\mathcal{L}_\text{br}=|{G}(\hat{y_0})-{G}(y)|_1,
\end{equation} 
where $G(\cdot)$ means the gray-scale version of an RGB image.

\paragraph{\textbf{Angular Color Loss}}
Increasing the brightness can amplify the color distortion from low-light images. We adopt a color loss~\cite{wang2019underexposed} that encourages the color of the enhanced images $\hat{y_0}$ to match the ground truth $y$. The color loss can be expressed as:
\begin{equation}
\mathcal{L}_\text{col}=\sum_{i}{\angle\left(\hat{y_0}_i,y_i\right)},
\end{equation} 
where $i$ denotes a pixel location and $\angle(,)$ calculates the angular difference between two 3-dimensional vectors representing colors in RGB color space.

\paragraph{\textbf{SSIM Loss}} We further incorporate 
Structural Similarity Index (SSIM) loss to improve the visual quality of the enhanced images.
SSIM calculates the structural similarity between a predicted image and a ground truth image, providing statistics for overall contrast and luminance consistency.
SSIM loss can be expressed as follows,
\begin{equation}
\mathcal{L}_\text{ssim} = \frac{(2\mu_{y}\mu_{\hat{y_0}} + c_1)(2\sigma_{y\hat{y_0}} + c_2)}{(\mu_{y^2} + \mu_{\hat{y_0}}^2 + c_1)(\sigma_{y^2} + \sigma_{\hat{y_0}}^2 + c_2)},
\end{equation} 
where $\mu_y$ and $\mu_{\hat{y_0}}$ are pixel value averages, $\sigma_y$ and $\sigma_{\hat{y_0}}$ are variances, $\sigma_{y\hat{y_0}}$ is covariance, $c_1$ and $c_2$ are constants for numerical stability.

\paragraph{\textbf{Perceptual Loss}}
Although effective, the pixel-space metrics mentioned above primarily focus on low-level information and are less aligned with human perception.
Thus, it is preferable to utilize high-level feature statistics to measure the quality of the enhanced images~\cite{zhang2018unreasonable}.
LPIPS~\cite{zhang2018perceptual} loss extracts deep features via a VGG network~\cite{simonyan2014very} pre-trained on ImageNet~\cite{deng2009imagenet} and calculates the distance between the enhanced features and ground truth features.
\begin{equation}
\mathcal{L}_\text{lpips}=\sum_l \frac{1}{H_lW_l}\vert \phi^l_{VGG}(\hat{y_0})-\phi^l_{VGG}(y)\vert_2,
\end{equation} 
where  $\phi^l_{VGG}$ represents the feature maps extracted from $l$-{th} layer of VGG network and $H_l,W_l$ are the height and weight of feature map of layer $l$, respectively. We find that LPIPS loss can improve the model's ability to restore high-frequency information.

Overall, the auxiliary loss functions can be summarized as:
\begin{equation}
\begin{aligned}
\mathcal{L}_\text{aux}=\mathcal{L}_\text{simple}+W_\text{br}\mathcal{L}_\text{br}+W_\text{col}\mathcal{L}_\text{col}+ W_\text{ssim}\mathcal{L}_\text{ssim}+W_\text{lpips}\mathcal{L}_\text{lpips},
\end{aligned}
\end{equation} 
where the $W_\text{br}$,$W_\text{col}$,$W_\text{ssim}$,$W_\text{lpips}$ are weighting coefficients.

\section{Experiment}

\begin{figure}
\centering
\includegraphics[width=0.9\columnwidth,keepaspectratio]{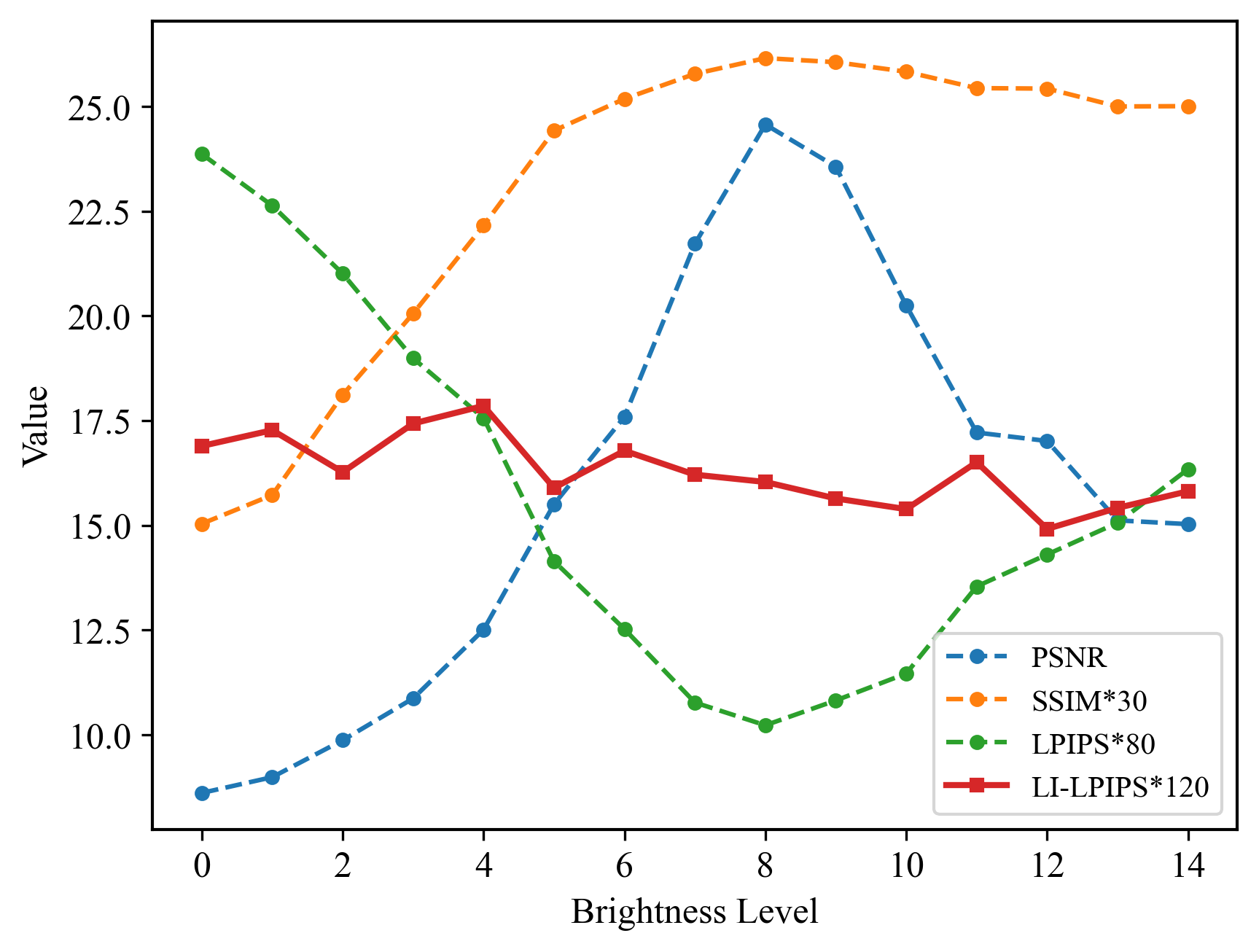}\\
\caption{The Metrics-Brightness Levels curves. We sample 15 images with different brightness levels as in Fig.~\ref{fig:over_control} and evaluate them via representative metrics. Both PSNR and SSIM exhibit inverted ``V''-shaped patterns, while LPIPS presents a ``V''-shaped pattern as brightness levels change.  
We introduce a novel metric, LI-LPIPS, which is more stable when brightness changes and better assesses image quality.}
\vspace{-3mm}
\label{fig:li-lpips}
\end{figure}

Due to the space limit, we provide more implementation details, results, and analysis in the supplementary.
\vspace{-3mm}

\subsection{Quantitative Metric}
Existing quantitative metrics typically assume the existence of an ideal brightness level, making it difficult to compare images with different brightness levels fairly. 
As shown in Fig.~\ref{fig:li-lpips}, PSNR and SSIM present an inverted ``V''-shaped pattern while LPIPS show a ``V''-shaped pattern.
These metrics exhibit large variances as brightness levels change, indicating their sensitivity to changes in light intensity.
As light enhancement is highly subjective, a better value of these metrics does not necessarily correlate with better image quality.
Additionally, the highly ill-posed nature of light enhancement brings about many possible optimal solutions. These solutions can have different white balances and exposure levels, further complicating the determination of a single "best" solution.

We are initially inspired by the color map's ability to normalize the brightness of images.
As mentioned in Eq.~\ref{eq:cloromap}, the color map can bring images with different brightness levels to a normalized brightness level, especially for images that are too dark or overexposed.
Then, we use Canny edge detector~\cite{canny1986computational} to extract useful structural information, further reducing the appearance information that is easily affected by illumination. 
We have empirically set the Canny threshold as 50 and 250. Other combinations exhibit minimal variation.
As semantic features are more resilient than pixel-space metrics, we select the LPIPS distance calculation for comparing the structure of two images. We name this new metric as Light-Independent LPIPS, dubbed $LI\text{-}LPIPS$:
\begin{equation}
\mathcal{L}_\text{LI\text{-}LPIPS}(a, b)=\sum_l \frac{1}{H_lW_l}\vert \phi^l(\text{Canny}(C(a))-\phi^l(\text{Canny}(C(b))\vert_2,
\end{equation}
where $C(\cdot)$ is color map, $\phi^l$ is the $l$-th layer feature map of a pre-trained VGG network, and $\text{Canny}(\cdot)$ represents the Canny operator. 
In Fig.\ref{fig:li-lpips}, when the average image brightness ranges from 0.18 to 0.69, the difference observed in LI-LPIPS is less than 0.02, while LPIPS yields a variation in maximum and minimum values as high as 0.15. This validates LI-LPIPS is insensitive to variations in brightness.

\begin{figure*}
\centering
\includegraphics[width=\textwidth,keepaspectratio]{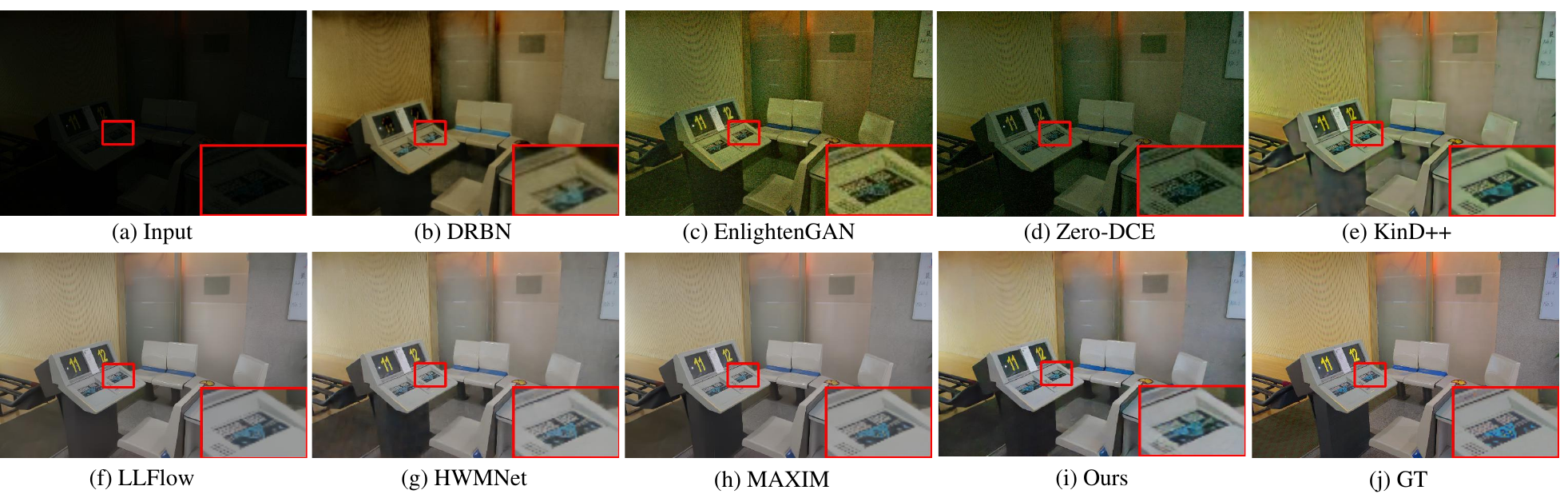}\\
\vspace{-2mm}
\caption{Results on LOL~\cite{Chen2018Retinex} test dataset. Our results exhibit fewer artifacts and are more consistent with the ground truth image.}
\label{fig:lol}
\end{figure*}

\begin{figure*}
\centering
\includegraphics[width=\textwidth,keepaspectratio]{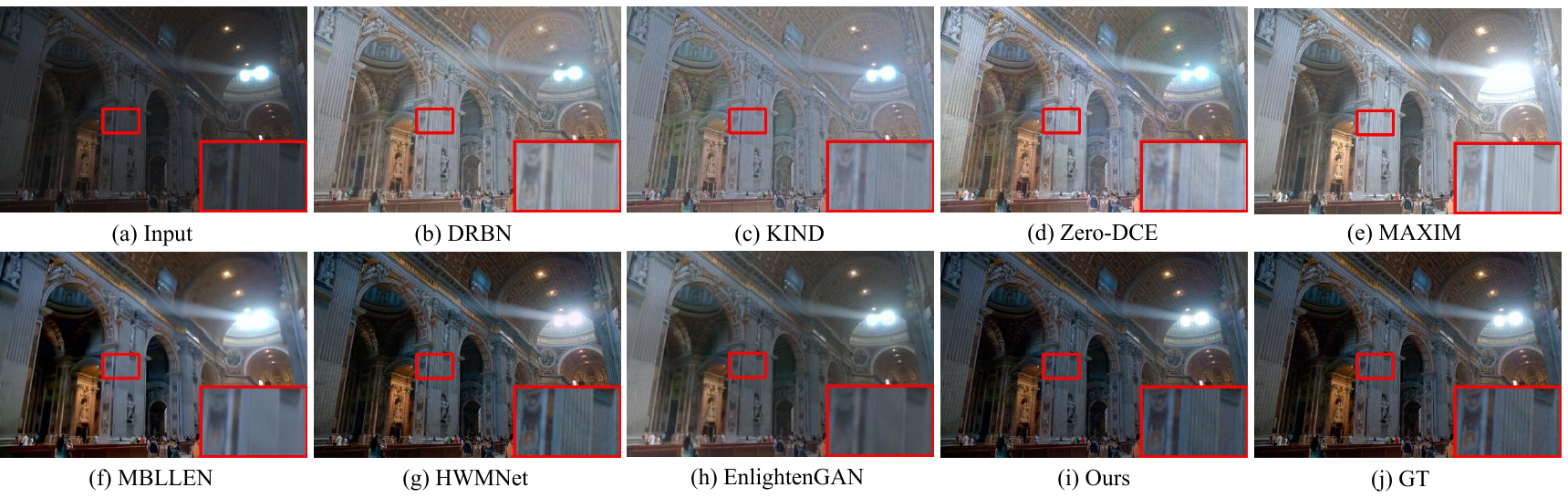}\\
\caption{Results on MIT-Adobe FiveK~\cite{fivek} test dataset. Our results exhibit less color distortion and contain richer details, which are more consistent with the ground truth.}
\label{fig:mit}
\end{figure*}

\subsection{Evaluation Protocol}
We evaluate the CLE-Diffusion on two popular benchmarks, LOL~\cite{Chen2018Retinex} and MIT-Adobe FiveK~\cite{fivek}. The LOL dataset contains 485 training and 15 testing paired images, with each pair comprising a low-light and a normal-light image.  
MIT-Adobe FiveK consists of 5000 images processed with Adobe Lightroom by five experts.
Following the split in previous methods~\cite{tu2022maxim,ni2020towards}, 4500 paired images are utilized as the training set, while the remaining 500 paired images serve as the test set.
We use PSNR, SSIM, LPIPS~\cite{zhang2018unreasonable}, and LI-LPIPS metrics to measure the quality of output images.

\begin{figure*}
\centering
\includegraphics[width=\textwidth,keepaspectratio]{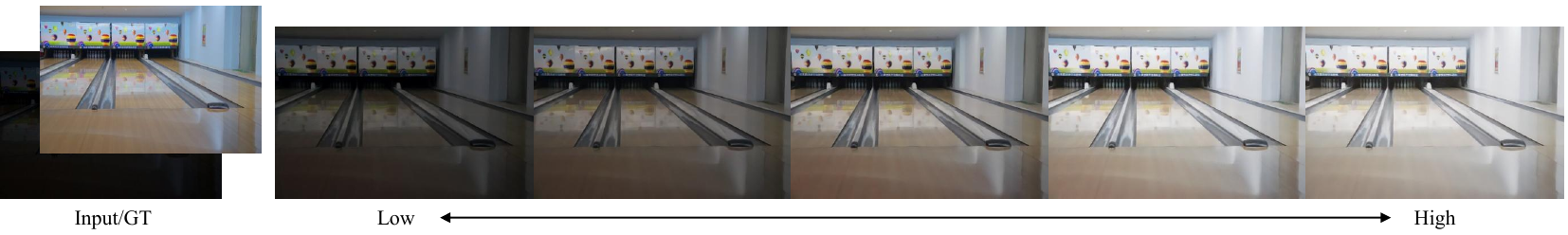}
\vspace{-4mm}
\caption{Visual results of global brightness control. By adjusting the brightness levels during inference, we can sample images with varying degrees of brightness while maintaining high image quality.}
\vspace{-3mm}
\label{fig:over_control}
\end{figure*}

\begin{figure*}
\centering
\includegraphics[width=\textwidth,keepaspectratio]{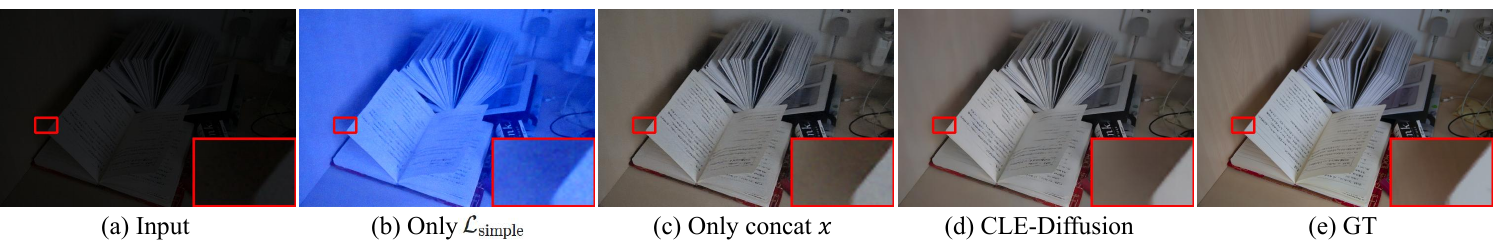}
\caption{Ablation study on the effectiveness of each component.}
\vspace{-3mm}
\label{fig:ablation}
\end{figure*}

\begin{table}
\centering
\caption{Comparisons on the LOL dataset. Ours* refers to the model variant trained with a smaller weight for color loss.}
\begin{tabular}{lcccc}
\toprule 
Method                 & PSNR↑ & SSIM↑ & LPIPS↓ & LI-LPIPS↓ \\
\midrule
Zero-DCE~\cite{guo2020zero}    & 14.86 & 0.54  & 0.33 & 0.3051   \\
EnlightenGAN~\cite{jiang2021enlightengan} & 17.48 & 0.65  & 0.32  &0.2838  \\
RetinexNet~\cite{Chen2018Retinex}  & 16.77 & 0.56  & 0.47 &0.5468   \\
DRBN~\cite{yang2020fidelity}    & 20.13 & 0.83  & \colorbox{yellow!25}{0.16} &0.3271   \\
KinD++~\cite{zhang2021beyond}   & 21.30  & 0.82  &\colorbox{yellow!25}{0.16}  &0.3768  \\
MAXIM~\cite{tu2022maxim}  & 23.43 & \colorbox{red!25}{0.96}  & 0.20  &\colorbox{yellow!25}{0.1801}      \\
HWMNet ~\cite{fan2022half}     &  24.24 & 0.85  & \colorbox{orange!25}{0.12}  &0.1893  \\
LLFlow ~\cite{wang2022low} &\colorbox{orange!25}{25.19}  &\colorbox{orange!25}{0.93}  &\colorbox{red!25}{0.11}  & \colorbox{orange!25}{0.1763}    \\
\textbf{Ours}        & \colorbox{red!25}{25.51} &  \colorbox{yellow!25}{0.89}  & \colorbox{yellow!25}{0.16}   & 0.1841  \\
\textbf{Ours*}        & \colorbox{yellow!25}{24.92} &  {0.88}  & \colorbox{yellow!25}{0.16}   & \colorbox{red!25}{0.1751}     \\
\bottomrule
\end{tabular}
\label{tab:lol}
\end{table}

\subsection{Performance on the LOL dataset}
We compare our CLE Diffusion with state-of-the-art low-light image enhancement methods. For quantitative comparison, we select several representative methods, including deep Retinex-based methods (RetinexNet~\cite{Chen2018Retinex},
KinD++~\cite{zhang2019kindling}),  CNN-based methods (MAXIM~\cite{tu2022maxim}, HWMNet~\cite{fan2022half}),
a Zero-Shot learning method (Zero-DCE~\cite{guo2020zero}), 
a Semi-Supervised Learning method (DRBN~\cite{yang2020fidelity}), 
a GAN-based (EnlightenGAN~\cite{jiang2021enlightengan})
and a Flow-based model (LLFlow ~\cite{wang2022low}). 

As shown in Tab.~\ref{tab:lol}, our method surpasses the compared methods in terms of PSNR and ranks third in SSIM, LPIPS, and LI-LPIPS. These findings provide evidence that our CLE Diffusion technique can generate superior samples from the dataset distribution in comparison to current approaches.

Visual comparisons are shown in Fig.~\ref{fig:lol}, 
our proposed method surpasses other methods by accurately aligning the predicted luminance with the ground truth.
The outputs generated by our framework exhibit significantly higher quality, characterized by well-suppressed noise, in contrast to the results produced by other methods. These alternative approaches display color shifts, noisy artifacts, and irregular illumination.

\subsection{Performance on MIT-Adobe FiveK dataset}
To further validate the ability of CLE Diffusion, we test on the MIT-Adobe FiveK dataset, which is much larger than LOL dataset and includes more diverse scenes. Tab.~\ref{tab:mit} shows the comparison results with other methods. 
Our method achieves the highest values in terms of PSNR, while also producing results comparable to state-of-the-art methods in terms of SSIM.
As indicated by Fig.~\ref{fig:mit}, our method is more consistent with ground truth in terms of color distortion. 
Compared to previous methods, our methods can preserve better details and color consistency as exemplified in zoomed-in regions.

\begin{table}
\centering
\caption{Comparisons on the MIT-Adobe FiveK dataset.}
\vspace{-3mm}
\setlength{\tabcolsep}{3mm}
\begin{tabular}{lcc}
\toprule 
Method       & PSNR↑ & SSIM↑   \\
\midrule
EnlightenGAN~\cite{jiang2021enlightengan} & 17.74 & 0.83    \\
CycleGAN ~\cite{zhu2017unpaired}  & 18.23 & 0.84    \\
Exposure~\cite{hu2018exposure} & 22.35 & 0.86    \\
DPE ~\cite{chen2018deep}        & 24.08 & 0.92    \\
UEGAN ~\cite{ni2020towards}        &  25.00 & {0.93}  \\
MAXIM ~\cite{tu2022maxim}      & \colorbox{yellow!25}{26.15} & \colorbox{yellow!25}{0.95}  \\
HWMNet ~\cite{fan2022half}        & \colorbox{orange!25}{26.29} & \colorbox{orange!25}{0.96}     \\

\textbf{Ours}         &  \colorbox{red!25}{29.81} &\colorbox{red!25}{ 0.97}   \\
\bottomrule
\end{tabular}
\label{tab:mit}
\vspace{-3mm}
\end{table}

\subsection{Controllable Light Enhancement}

With the powerful capabilities of CLE-Diffusion and Mask-CLE-Diffusion, we can not only control the global brightness of images at a specified brightness level but also achieve precise brightness control for desired regions.
Users can input multiple desired brightness levels into the network to sample multiple images with various brightness levels as depicted in Fig.~\ref{fig:over_control}.
For images with distinct brightness levels, the fidelity of details is remarkably high and the images exhibit no apparent color distortion, overexposure, or underexposure while maintaining a consistent global brightness. 

By utilizing SAM model~\cite{sam}, users can effortlessly obtain regions of interest(ROI) by clicking on the image. As demonstrated in Fig.~\ref{fig:region_control}, Mask-CLE-Diffusion is able to selectively enhance specific regions to attain desired brightness levels.
In contrast to directly blending normal-light and low-light images using masks, our results have a more natural appearance and better emphasize the ROI.

\subsection{Ablation Study}
We conduct ablation studies to validate the effectiveness of our proposed loss functions and network design.
As shown in Fig.~\ref{fig:ablation}, when training with only $\mathcal{L}_\text{simple}$, the output images suffer from color distortion, artifacts, and unsatisfactory lighting. And when training without the conditioning color map and SNR map, the network produces noisy results which do not appear in the results of the full model.
Tab.~\ref{tab:ablation} also presents a notable decrease in all four metrics.
These results show that auxiliary losses, the conditioning color map, and the SNR map contribute to improving the overall performance of the model.

\begin{table}
\centering
\caption{ Ablation study for CLE Diffusion's components.}
\vspace{-3mm}
\begin{tabular}{lcccc}
\toprule 
Method                 & PSNR↑ & SSIM↑ & LPIPS↓ & LI-LPIPS↓ \\
\midrule
 Only $\mathcal{L}_\text{simple}$       &8.85   &0.59  &0.63   &0.2940         \\
Only concat $x$      &22.22   &0.81   &0.22   &0.2186          \\
CLE-Diffusion       & 25.51 &  0.89  & 0.16   & 0.1841     \\
\bottomrule
\end{tabular}
\label{tab:ablation} 
\vspace{-3mm}
\end{table}

\section{Conclusion}
In this work, we present a novel diffusion framework, CLE Diffusion, for Controllable Light Enhancement.
The framework is based on a diffusion model conditioned on an illumination embedding, which enables seamless control of brightness during inference.
Additionally, we incorporate the Segment-Anything Model (SAM) to allow users to easily enhance specific regions of interest with a single click. Through extensive experiments, we demonstrate that our CLE Diffusion achieves competitive results in terms of quantitative metrics, qualitative performance, and versatile controllability.

\noindent{\textbf{Limitations.}}
Despite the exciting results from CLE Diffusion, the model suffers from the slow inference speed of diffusion models, since multiple inferences are required during the sampling process. Moreover, challenging cases (e.g., complex lighting, blurry scenes) need to be further explored. In the future, we will investigate how to extend our framework to more general scenarios.

\noindent{\textbf{Acknowledgment.}}
This work was supported in part by the Fundamental Research Funds for the Central Universities (No.K22RC00010) and A*STAR Career Development Funding ({CDF})
Award (No.222D-
800031).

\bibliographystyle{ACM-Reference-Format}
\bibliography{egbib}

\clearpage
\appendix
\renewcommand{\thefigure}{\Alph{figure}}
\setcounter{figure}{0}
\renewcommand{\thetable}{\Alph{table}}
\setcounter{table}{0}

\section{Implementation Details}
All experiments are conducted on one NVIDIA RTX3090 GPU with PyTorch. The Adamw optimizer is used with the initial learning rate of $5 \times 10^5$. The weight decay is $1 \times 10^4$. A batch size of 16 is applied for 12000 epochs in LOL dataset~\cite{Chen2018Retinex} and 2,000 epochs in MIT-Adobe FiveK dataset~\cite{fivek}.  The brightness level information is derived from the average values of the patches. We set $W_{br}$,$W_{col}$,$W_{ssim}$,$W_{lpips}$  to 20,100,2.83 and 50, respectively.

\begin{figure*}[b]
\centering
\includegraphics[width=\textwidth,keepaspectratio]{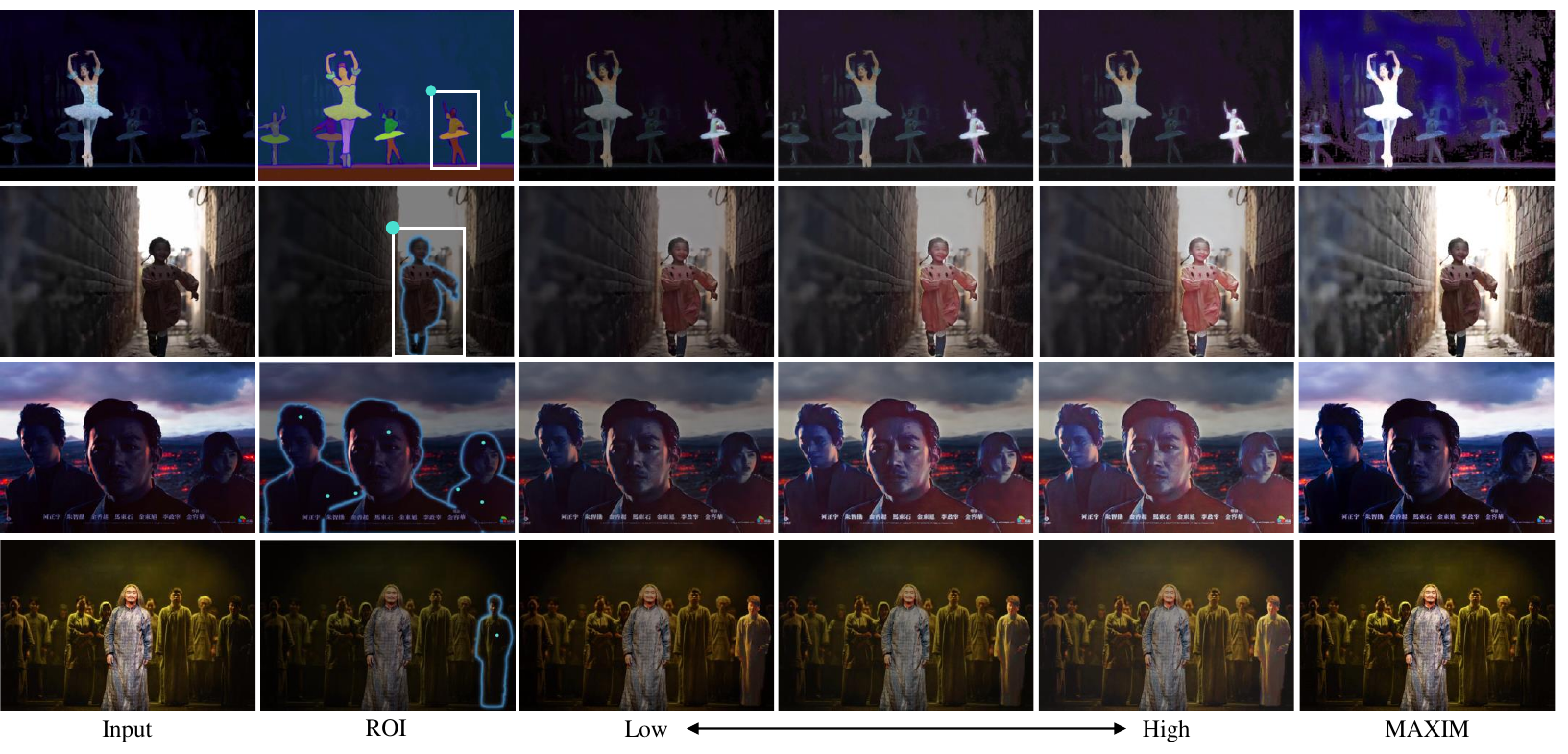}\\
\caption{More cases about region controllable light enhancement. Equipped with the Segment-Anything Model (SAM), users can designate regions of interest (ROI) using simple inputs like points or boxes. Our model facilitates controllable light enhancement within these regions, producing results that blend naturally and seamlessly with the surrounding environment.}
\label{fig:region control}
\end{figure*}

\begin{figure*}
\centering
\includegraphics[width=\textwidth,keepaspectratio]{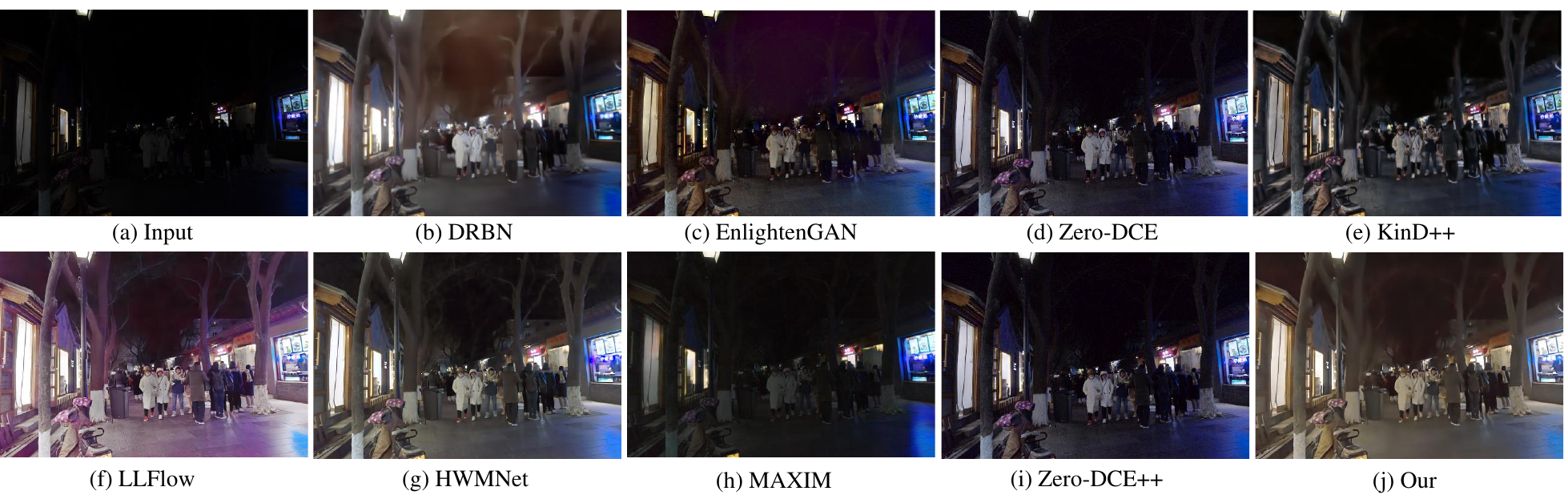}\\
\caption{Comparisons on a real-world image from VE-LOL dataset~\cite{liu2021benchmarking}. Other methods often rely on well-lit brightness extracted from pre-existing datasets, limiting their applicability to diverse scenarios. Unlike most methods that struggle to enhance the brightness at night sufficiently, our method incorporates a brightness control module, allowing us to sample images with higher brightness that appear more natural in such situations.  }
\label{fig:velol}
\end{figure*}

During the training phase, we randomly crop images into 128 $\times$ 128 randomly and perform horizontal and vertical flipping for data augmentation. 
Similar to prior works, we normalize the input pixels into the range of $[-1, 1]$ to stabilize the training.
However, the datasets for low-light image enhancement are primarily distributed in the lower range of values and exhibit characteristics of a short-tailed distribution. 
We empirically discover that normalizing the input data to Gaussian distribution can ease the optimization.

\section{More Quantitative comparisons}
As shown in Tab.~\ref{tab:mit2}, on the MIT-Adobe FiveK dataset, we outperform state-of-the-art methods and achieve the best performance across all metrics. To ensure consistency, we revisited our testing scheme and adopted the procedure described in MAXIM~\cite{tu2022maxim}, which involves center cropping the images to 512x512 prior to metric calculations. Additionally, we identified that in MAXIM~\cite{tu2022maxim}, the reported PSNR and SSIM values are computed solely on the Y channel (grayscale version). Consequently, we recalculated our performance on the MIT-Adobe FiveK dataset and showcased our superiority compared to prior methods, achieving a notable improvement of +3.65 dB.

\begin{table}[htbp]
    \centering
    \caption{Quantitative comparisons on MIT-Adobe Fivek dataset}
    \label{tab:comparison}
    \begin{tabular}{p{0.9cm} p{0.7cm} p{0.7cm} p{0.7cm} p{0.7cm} p{0.7cm} p{1.3cm} }
        \hline
        & PSNR$\uparrow$ & PSNR-Gray$\uparrow$ & SSIM$\uparrow$ & SSIM-Gray$\uparrow$ & LPIPS$\downarrow$ & LI-LPIPS$\downarrow$ \\
        \hline

        LLFlow & 19.74 & 21.04 & 0.80 & 0.80 & 0.1728 & 0.2665 \\
        HWMNet & 24.41  &26.30  & 0.93  & 0.96  & 0.0812  &0.1428 \\
        MAXIM & 24.60 & 26.16 & 0.93 & 0.95 & 0.0752 & 0.1491 \\
        \textbf{Our} & \textbf{26.62} & \textbf{29.81} & \textbf{0.94} & \textbf{0.97} & \textbf{0.0601} & \textbf{0.1385} \\
        \hline
    \end{tabular}
    \label{tab:mit2}
\end{table}

\section{More Ablation Study}
To further explore the contribution of each auxiliary loss function, we present the results of CLE Diffusion trained with various combinations of losses in Tab.~\ref{tab:abl}. Removing the brightness loss fails to achieve sufficient brightness enhancement. The angular color loss function contributes to reducing color distortion and improves overall performance. Furthermore, we separately train the networks without SSIM loss or perceptual loss to demonstrate the positive impact of the auxiliary loss functions.

\begin{table}[ht]
\centering
\caption{ Quantitative comparisons on the LOL dataset for ablations of each auxiliary loss function. The full model presents better results than its partial versions, demonstrating the effectiveness of the auxiliary loss functions.}
\begin{tabular}{p{0.6cm} p{0.4cm} p{0.4cm} p{0.4cm} p{0.4cm} p{0.6cm} p{0.6cm} p{0.6cm} p{1.3cm}}
\hline
$L_{\text{simple}}$ & $L_{\text{br}}$ & $L_{\text{col}}$ & $L_{\text{ssim}}$ & $L_{\text{lpips}}$ & PSNR$\uparrow$ & SSIM$\uparrow$ & LPIPS$\downarrow$ & LI-LPIPS$\downarrow$ \\
\hline
$\checkmark$ & & & & & 8.85 & 0.59 & 0.63 & 0.2940 \\
$\checkmark$ & & $\checkmark$ & $\checkmark$ & $\checkmark$ & 22.45 & 0.85 & 0.20 & 0.1908 \\
$\checkmark$ & $\checkmark$ & & $\checkmark$ & $\checkmark$ & 25.15 & 0.87 & 0.17 & 0.1885 \\
$\checkmark$ & $\checkmark$ & $\checkmark$ & & $\checkmark$ & 25.04 & 0.87 & 0.18 & 0.1866 \\
$\checkmark$ & $\checkmark$ & $\checkmark$ & $\checkmark$ & & 25.07 & 0.86 & 0.20 & 0.1918 \\
$\checkmark$ & $\checkmark$ & $\checkmark$ & $\checkmark$ & $\checkmark$ & 25.51 & 0.89 & 0.16 & 0.1841 \\
\hline
\label{tab:abl}
\end{tabular}

\end{table}

\section{More visual comparisons}

\begin{figure*}
\centering
\includegraphics[width=\textwidth,keepaspectratio]{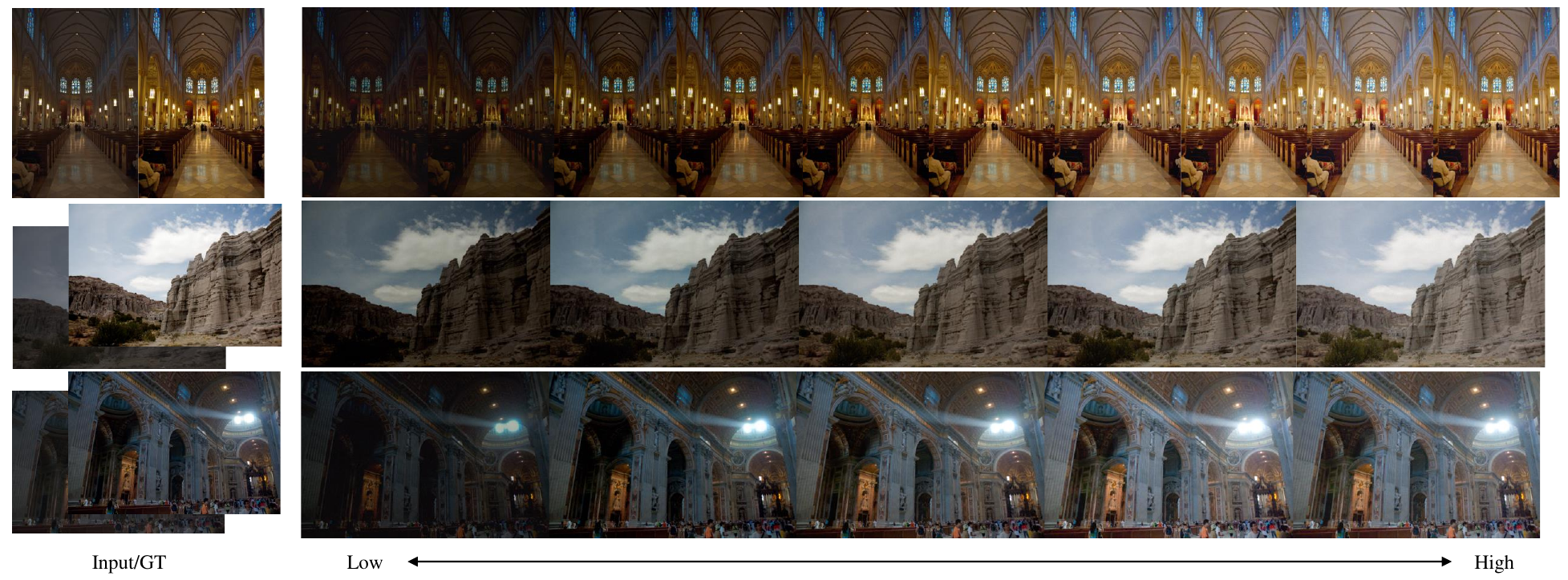}\\
\caption{Global Controllable Light Enhancement on MIT-Adobe FiveK dataset. Our method enables users to select various brightness levels, even significantly brighter than the ground truth.}
\label{fig:mit global control}
\end{figure*}

\begin{figure*}
\centering
\includegraphics[width=\textwidth,keepaspectratio]{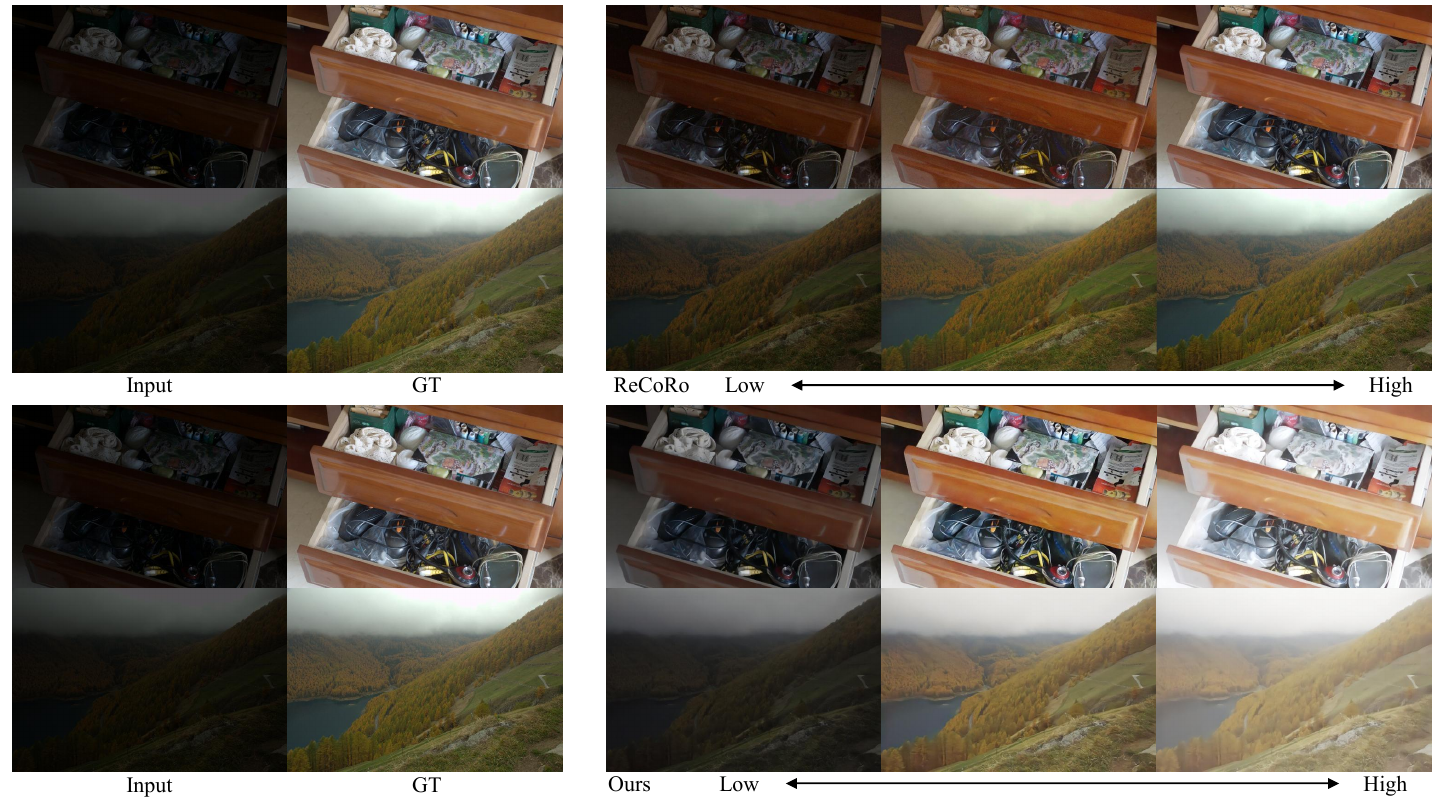}\\
\caption{Global brightness control compared with ReCoRo~\cite{xu2022recoro}. While ReCoRo is constrained to enhancing images with brightness levels that fall between low-light and ``well-lit'' images, our model can handle a wider range of brightness levels. It can be adjusted to sample any desired brightness, providing greater flexibility and control over different lighting conditions.
}
\label{fig:recoro}
\end{figure*}

\begin{figure*}
\centering
\includegraphics[width=\textwidth,keepaspectratio]{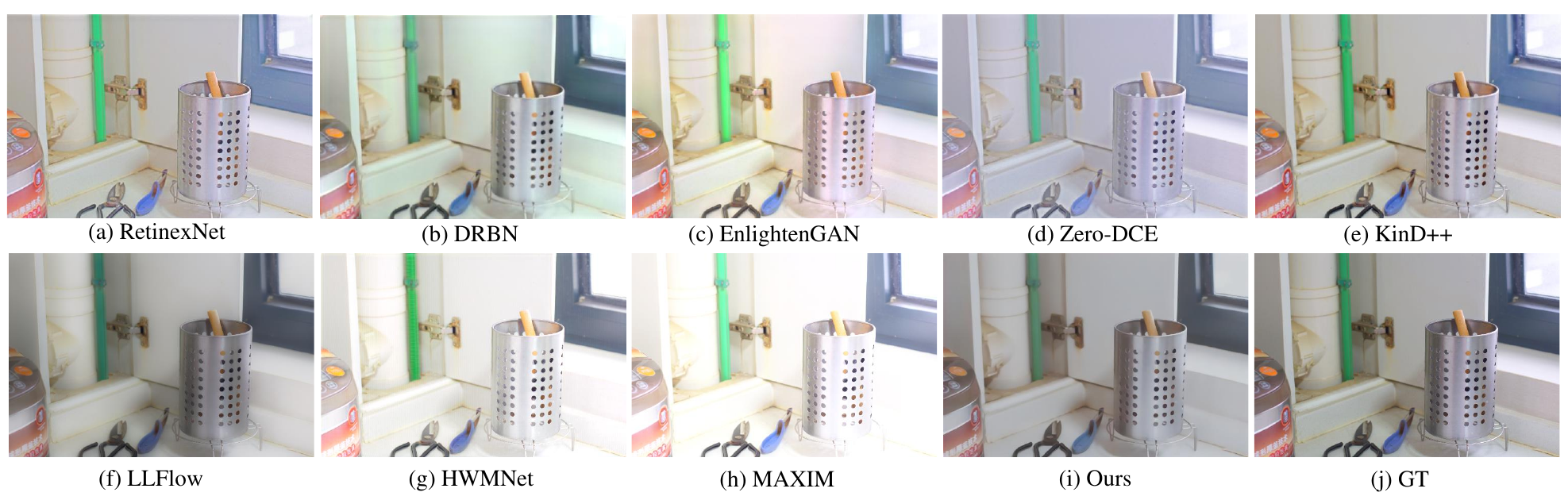}\\
\caption{Performance on normal light image inputs. We utilize the normal-light images from the LOL dataset as inputs to evaluate the models' capability in handling high-light images. HWMNet and MAXIM exhibit overexposure in certain regions, resulting in considerably over-exposed images. LLFlow produces blurred images, while other methods result in color distortion. Our method achieves visually pleasing results in terms of color and brightness.}
\label{fig:overexposure}
\end{figure*}

\begin{figure*}
\centering
\includegraphics[width=\textwidth,keepaspectratio]{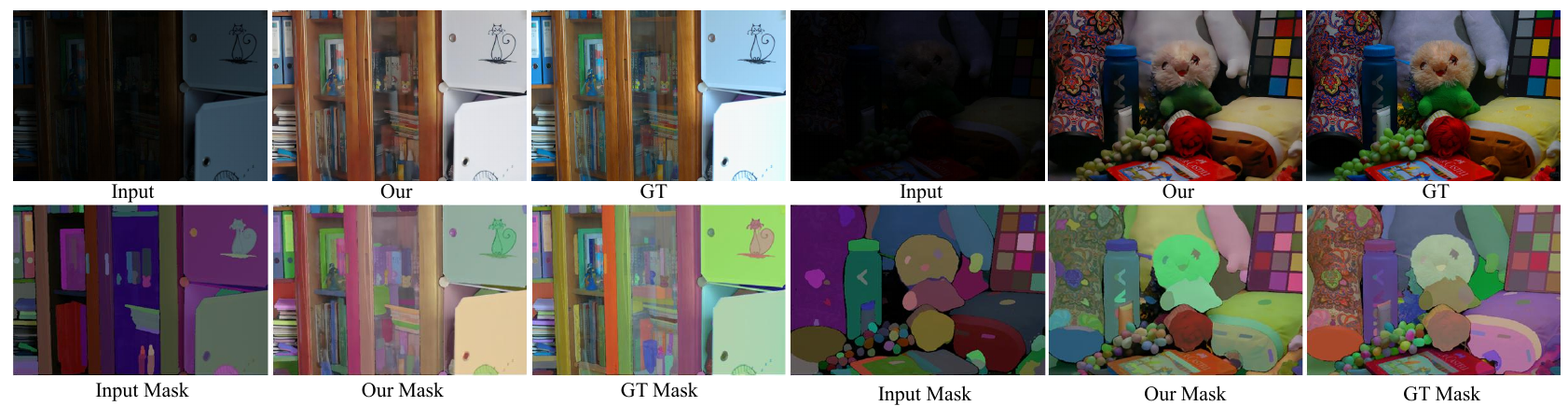}\\
\caption{Performance on Segment-Anything model. SAM generates coarse masks when dealing with images captured in low-light conditions. After enhancing with our model, the images can be effectively segmented even in dark environments. This demonstrates our model's ability to restore details that are friendly to high-level machine vision models.}
\label{fig:sam}
\end{figure*}

\begin{figure*}
\centering
\includegraphics[width=\textwidth,keepaspectratio]{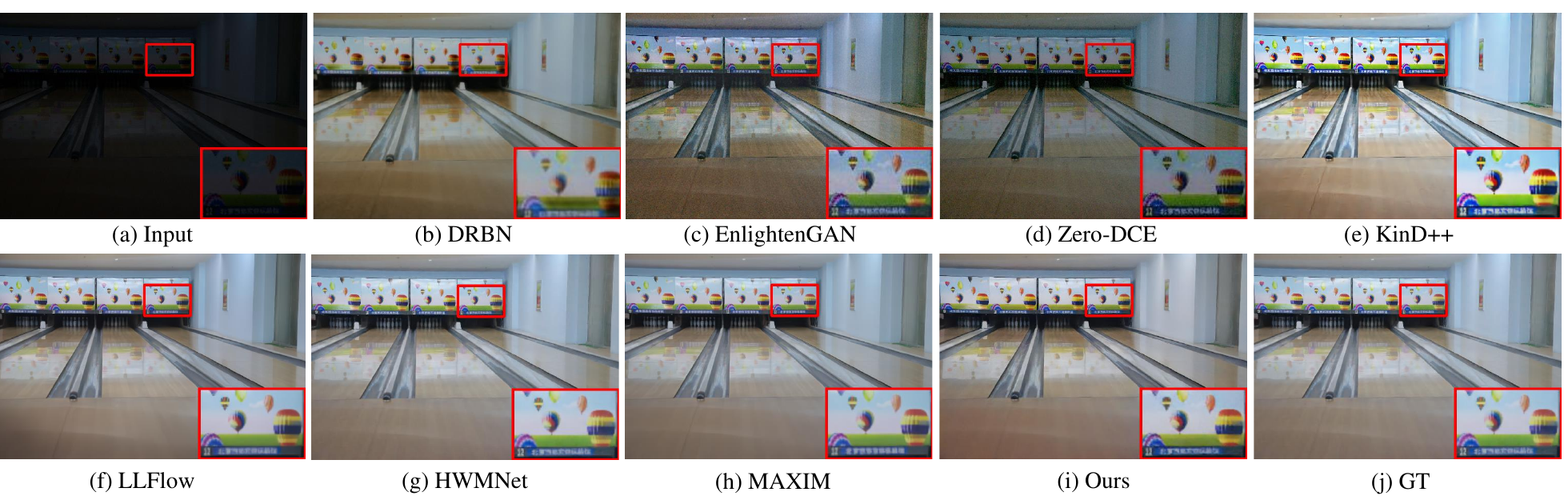}\\
\caption{More comparisons on LOL dataset.  }
\label{fig:lol2}
\end{figure*}

We provide more cases about region controllable light enhancement in Fig.~\ref{fig:region control}.
Fig.~\ref{fig:velol} show comparisons of performance on a real-world image.
Fig.\ref{fig:mit global control} shows global controllable light enhancement on MIT-Adobe FiveK dataset.
We provide comparisons on global brightness control ability in Fig.\ref{fig:recoro}.
We also compare performance on normal light image inputs in Fig.~\ref{fig:overexposure}.
Fig.~\ref{fig:sam}  shows comparisons of performance on Segment-Anything model.
Fig.~\ref{fig:lol2} give comparisons on LOL dataset.

\end{document}